\documentclass{article}

\usepackage[final,main]{agents4science_2025}

\usepackage[utf8]{inputenc} 
\usepackage[T1]{fontenc}    
\usepackage{hyperref}       
\usepackage{url}            
\usepackage{booktabs}       
\usepackage{amsfonts}       
\usepackage{nicefrac}       
\usepackage{microtype}      
\usepackage{xcolor}         
\usepackage[table]{xcolor}
\usepackage{graphicx}   
\usepackage{colortbl}   
\usepackage{subcaption}
\usepackage{amsmath}
\usepackage{wrapfig}
\usepackage{algorithm}
\usepackage{algpseudocode}
\usepackage{natbib}

\newcommand{\modelname}{TEAM-PHI}

\title{Towards Automatic Evaluation and Selection of \\ PHI De-identification Models\\ via Multi-Agent Collaboration}

\author{%
  Guanchen Wu$^{1}$, Zuhui Chen$^{2}$, Yuzhang Xie$^{1}$, Carl Yang$^{1}$ \\
  $^{1}$Department of Computer Science, Emory University \\
  $^{2}$Department of Statistics, Columbia University \\
  \texttt{\{guanchen.wu, yuzhang.xie, j.carlyang\}@emory.edu} \\
  \texttt{zc2748@columbia.edu} \\
}

\begin{document}

\maketitle

\begin{abstract}

Protected health information (PHI) de-identification is critical for enabling the safe reuse of clinical notes, yet evaluating and comparing PHI de-identification models typically depends on costly, small-scale expert annotations. We present \modelname, a \emph{multi-agent evaluation and selection framework} that uses large language models (LLMs) to automatically measure de-identification quality and select the best-performing model without heavy reliance on gold labels.  
\modelname\ deploys multiple \textit{Evaluation Agents}, each independently judging the correctness of PHI extractions and outputting structured metrics.  
Their results are then consolidated through an LLM-based majority voting mechanism that integrates diverse evaluator perspectives into a single, stable, and reproducible ranking. Experiments on a real-world clinical note corpus demonstrate that \modelname\ produces consistent and accurate rankings: despite variation across individual evaluators, LLM-based voting reliably converges on the same top-performing systems.  
Further comparison with ground-truth annotations and human evaluation confirms that the framework’s automated rankings closely match supervised evaluation. By combining independent evaluation agents with LLM majority voting, \modelname\ offers a practical, secure, and cost-effective solution for \emph{automatic evaluation and best-model selection} in PHI de-identification, even when ground-truth labels are limited. \footnote{The implementation details, prompt designs, and codes are available in the anonymous repository at \href{https://anonymous.4open.science/r/SAFE-0915}{https://anonymous.4open.science/r/SAFE-0915}.}

\end{abstract}

\section{Introduction}
Clinical notes—such as discharge summaries, nursing notes, and radiology or ECG reports—capture rich context like clinical reasoning, observations, and social factors that are often missing in structured electronic health records \citep{seinen2025using, tayefi2021challenges, zhang2024tacco}. Despite their value for research and downstream applications, clinical notes often contain protected health information (PHI), including names, dates, and addresses, and their reuse is strictly regulated under privacy laws such as HIPAA \citep{moore2019review, cohen2018hipaa}. To enable safe sharing and analysis, PHI de-identification is typically applied as a preprocessing step, automatically detecting and removing or replacing personal identifiers—for example, transforming ``John Smith was admitted on 03/15/2024 at Hospital A'' into ``[NAME] was admitted on [DATE] at [HOSPITAL]''.

Existing PHI de-identification (De-id) systems have progressed from handcrafted rules and feature-based machine learning to deep learning and transformer-based models \citep{kovavcevic2024identification}. More recently, large language models (LLMs) have opened new avenues for clinical NLP tasks such as named entity recognition, relation extraction, and note summarization \citep{lehman2023we, wu2024ontology}. Studies show that multiple kinds of LLMs can effectively extract PHI entities, generate structured representations from free text, and assist in clinical decision-making \citep{wu2025large, liu2023deid}

Although LLM-based PHI de-identification has significantly advanced the field, methods for systematically evaluating and automatically selecting the best de-identification model remain underexplored. Conventional evaluation relies on costly, small, institution-specific expert annotations, which impede the creation of reliable, generalizable benchmarks \citep{altalla2025evaluating}. LLMs have shown broad prior knowledge and strong zero-shot capabilities \citep{chang2024survey}, offering a promising route to assess de-identification models without gold labels. However, using LLMs as verifiers raises new challenges: how to design efficient, calibrated, and robust evaluation pipelines that are insensitive to prompt phrasing or choice of verifier, and how to ensure their judgments are trustworthy across use cases \citep{chang2024survey,de2025study}.

In this paper, we propose \modelname\   (\underline{T}rusted \underline{E}valuation and \underline{A}utomatic \underline{M}odel selection for \underline{PHI}), a multi-agent framework to automatically evaluate LLM-based de-identification methods and select the best-performing model. \modelname\ first runs multiple LLMs as De-id models that extract PHI from raw clinical text in a structured format. Multiple evaluation agents then assess those De-id outputs without relying on gold annotations. We validate the framework on a real-world clinical-note corpus and find that (1) aggregating Evaluation Agents by majority voting improves reliability relative to single-model assessments, (2) the multi-agent evaluators operate efficiently and produce consistent relative rankings, and (3) Llama-70B consistently emerges as the most reliable De-id model — a result supported by ablation studies using a limited manually annotated test set and independent human expert review.

Our contributions are threefold: (i) we introduce \modelname, a multi-agent framework that uses multiple LLMs to perform automated PHI evaluation and select the best de-identification model without heavy reliance on human labels; (ii) we demonstrate through extensive experiments that the multi-agent design yields consistent and accurate estimates of de-identification quality and that voting aggregation improves evaluator reliability; and (iii) we show that \modelname\ is a practical, secure, and cost-effective tool for guiding deployment of privacy-preserving data pipelines in healthcare settings.

\section{Related Work}

\paragraph{PHI De-identification.} De-identification of PHI has evolved from rule-based systems with handcrafted dictionaries and regexes \citep{uzuner2007evaluating, meystre2010automatic} to statistical learners (CRFs, SVMs) \citep{he2015crfs, jiang2017identification}, then to deep models (BiLSTM-CRF) \citep{dernoncourt2017identification, tang2020identification} and transformer-based architectures (e.g., BioBERT) \citep{lee2020biobert, johnson2020deidentification}, improving accuracy while reducing manual feature engineering. Recently, LLMs have been leveraged for many clinical tasks \citep{xie2024promptlink, wu2024ontology}, including PHI de-identification, because of their strong contextual reasoning and zero-/few-shot learning abilities \citep{bhasuran2025preliminary}. 
Recent work has demonstrated the promise of large language models (LLMs) for PHI De-id. For example, DeID-GPT leverages GPT-4 for zero-shot de-identification of medical text, achieving competitive performance without task-specific fine-tuning \citep{liu2023deid}. Similarly, \citet{wu2025large} proposed the LPPA framework, which combines synthetic data generation with instruction tuning to fine-tune local LLMs for PHI extraction. LPPA offers a privacy-preserving solution while notably improving recall—a critical metric for safeguarding sensitive health data—and its hybrid models (LPPA-4K and LPPA-5K) are also adopted in our implementation of De-id models. Despite these advances, existing systems still lack robust methods for systematically evaluating and automatically selecting the best De-id model, motivating the development of our proposed framework.

\paragraph{Evaluation of De-identification Systems.} 
Traditionally, PHI de-identification systems are evaluated against gold-standard corpora such as the i2b2 shared tasks \citep{stubbs2015automated}. Metrics typically include precision, recall, and F1-score. However, such datasets are costly to create and limited in scope. Alternative evaluation approaches include weak supervision, ensemble frameworks, and extensible benchmarking systems designed to assess de-identification tools across multiple corpora and settings \citep{heider2024extensible}. Despite recent progress, robust evaluation without gold-standard annotations, as in this study, is still largely unexplored.

\paragraph{Large Language Models as Judges.} 
Some researchers has begun to examine the use of LLMs as evaluators, or ``judges,'' in a range of NLP tasks including summarization, dialogue, and machine translation. These studies show that LLMs can approximate human judgments with promising consistency \citep{zhu2023judgelm, hada2023large, xie2025kerap, pan2024graphnarrator}. Specialized evaluator models such as Prometheus incorporate rubric-based scoring to align more closely with human ratings \citep{kim2023prometheus}. However, concerns of bias, calibration, and fairness remain when deploying LLM-based judges at scale \citep{wang2023large}. Our work extends this line of research to the sensitive domain of PHI de-identification, systematically evaluating whether Evaluation Agents can replicate gold-standard and human assessments when ground truth is limited or masked.

\section{Method}

\subsection{Problem Formulation}

\paragraph{Notation.}  
All notations used throughout the paper are summarized in Appendix~\ref{app:notation}.

\paragraph{Task Definition.}  
Let $\mathcal{X}$ denote the space of clinical notes, and let $\mathcal{C} = \{c_1, \ldots, c_m\}$ be the finite set of PHI categories (e.g., \textsc{PERSON}, \textsc{DATE/TIME}, \textsc{LOCATION}, \textsc{ORGANIZATION}).  
For a given note $x \in \mathcal{X}$, the de-identification output is a set of labeled entity spans:
\begin{equation}
\label{eqa:de-id}
\texttt{PHI}(x) = \{(c_i, e_i)\}_{i=1}^{k},
\end{equation}
where $c_i \in \mathcal{C}$ and $e_i \subseteq x$ is a contiguous text span corresponding to an entity of type $c_i$.  
Our objective is to select the best-performing de-identification model from a candidate set $\mathcal{M} = \{M_1, \ldots, M_R\}$ by maximizing a utility function that does not require gold-standard annotations:
\begin{equation}
\label{eqa:select}
M^* = \arg\max_{M \in \mathcal{M}} \ \mathbb{E}_{x \sim \mathcal{D}} \ \texttt{PHI-EV}(M, x),
\end{equation}
where $\texttt{PHI-EV}(M, x)$ denotes the evaluation score for model $M$ on input $x$.

\paragraph{Conventional Evaluation.}  
In traditional settings, given a model $M$, its prediction $\texttt{PHI}_M(x)$, and the corresponding gold annotations $y^*(x)$, the evaluation score is computed via a standard metric:
\begin{equation}
\label{eqa:conventional}
\texttt{PHI-EV}(M, x) = f\big(\texttt{PHI}(M, x),\ y^*(x)\big),
\end{equation}
where $f$ is typically precision, recall, or F1-score.

\paragraph{\modelname\ Evaluation.}  
Due to the scarcity of large-scale annotated corpora $\{(x, y^*(x))\}$ in real-world applications, we propose replacing $f$ with a multi-agent evaluation function $g$ that aggregates judgments from $A$ independent LLM-based evaluation agents $\{E_a\}_{a=1}^A$:
\begin{equation}
\label{eqa:multi-agent}
\texttt{PHI-EV}(M, x) = g\big(\texttt{PHI}(M, x)\big), \qquad g = \mathrm{Aggregate}(E_1, \ldots, E_A) \quad \text{(e.g., majority voting)}.
\end{equation}

\subsection{Framework Overview}

\begin{figure}[htbp]
\vspace{-10pt} 
\centering
\includegraphics[width=0.8\textwidth]{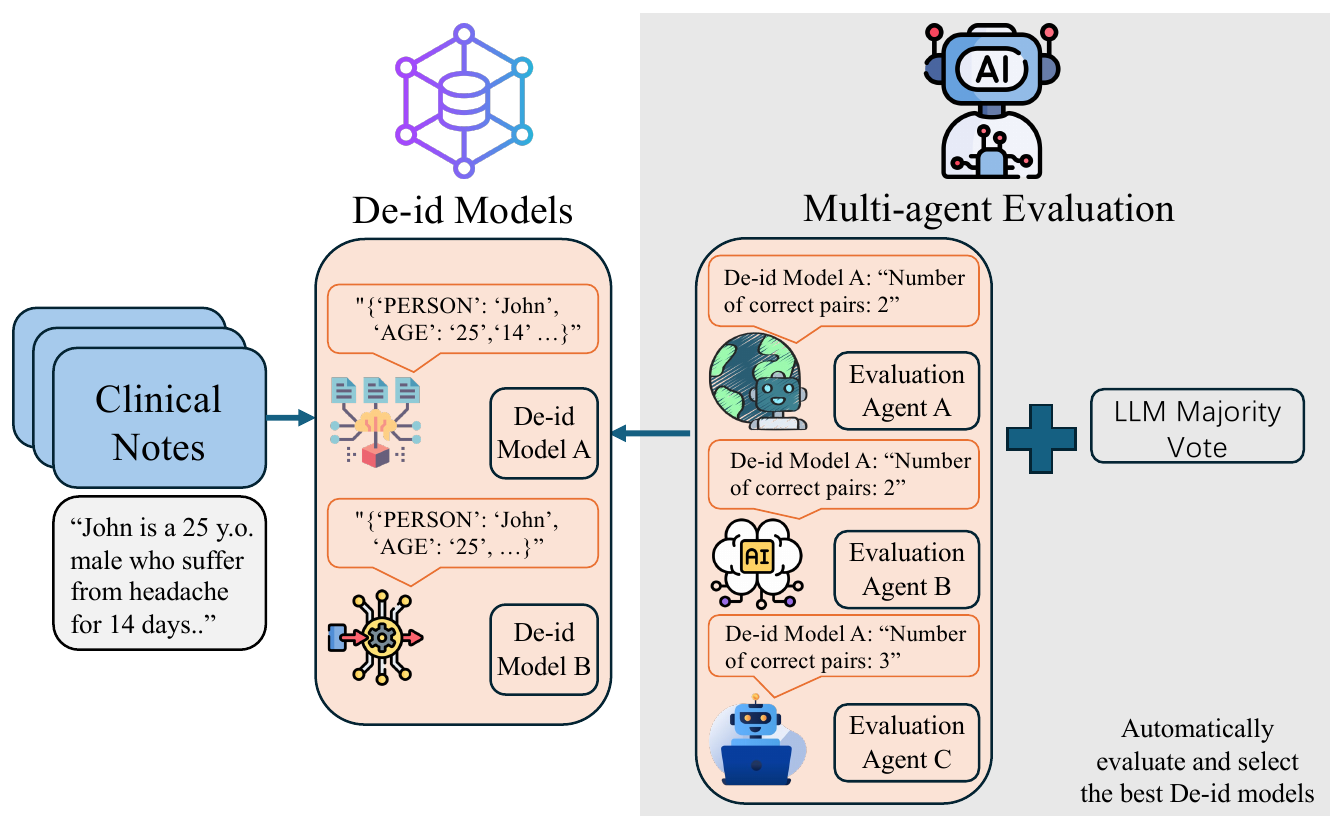}
\caption{Framework of \modelname.}
\label{fig:1}
\vspace{-10pt}
\end{figure}

To meet this objective, we propose a multi-agent evaluation framework \ref{fig:1} that decouples PHI extraction from quality assessment.  
First, raw clinical notes are processed in parallel by multiple De-id models, each producing a set of PHI predictions.  
These predictions are then evaluated by a pool of independent \textit{Evaluation Agents}—LLM-based judges that assess the correctness of every predicted PHI entity without relying on gold annotations.  
Their judgments are aggregated through majority voting and further verified by targeted human review to generate trustworthy model comparisons and rankings.  
This modular design supports large-scale benchmarking of de-identification systems without human-labeled corpora and can flexibly incorporate new de-identification or evaluation models without altering the overall protocol.  


\subsection{De-id Models}


We employ a diverse set of De-id models—eight LLMs, including two LPPA hybrids trained on synthetic clinical notes \citep{wu2025large}. Each model receives the same prompt and outputs predictions as a JSON dictionary of PHI entities, ensuring a unified, machine-readable format across heterogeneous architectures. Although these models differ in scale, training data, and internal reasoning, our framework treats them purely as black boxes and focuses on rigorous and reproducible evaluation and selection through multi-agent collaboration. The process are described in Eq.(\ref{eqa:conventional}).

\subsection{Multi-agent Evaluation}
\label{sec:multi-agent-evaluation}

Evaluation Agents serve as independent judges of the quality of PHI de-identification. Given a clinical note $x$ and the predicted output $\texttt{PHI} =\{ (c_1, e_1), \cdots, (c_i, e_i) \}$, the Evaluation Agents $\{E_a\}_{a=1}^A$ will evaluate De-id model outputs and select the best performing model follow Eq.(\ref{eqa:multi-agent}). Evaluation Agents face challenges distinct from those of De-id models, and our design addresses them explicitly.  

\paragraph{Normalization of linguistic variation.} PHI entities may appear in multiple surface forms, such as different name prefixes, diverse date formats, or varied expressions of age. To handle these cases we introduce a normalization procedure that standardizes entity spans (e.g., mapping ``Dr. Smith'' and ``Smith'' to the same \textsc{PERSON} mention, or converting ``03/15/2024'' and ``March 15, 2024'' to a canonical date).  

\paragraph{Consistent and machine-readable output.} To ensure reproducibility and facilitate automated scoring, each Evaluation Agent is constrained to output its decision as a strict JSON object of the form \texttt{\{"Number of Correct Pairs":N\}}, where $N$ is the integer total of validated predictions. This removes ambiguity caused by formatting differences and allows direct aggregation of results across agents.  

\paragraph{Mitigating evaluator bias.} While De-id models must identify all PHI entities, Evaluation Agents must critically verify predictions without over-accepting incorrect ones or missing valid but differently expressed entities. To reduce individual bias and quantify reliability, multiple Evaluation Agents are deployed in parallel, and their outputs are combined through ensemble strategies such as majority voting.

\paragraph{LLM Majority Voting.}
To consolidate the independent judgments from multiple Evaluation Agents into a single, reproducible ranking of De-id models, we introduce an LLM majority voting mechanism.  
This mechanism treats each Evaluation Agent’s summary of correctness counts as a “ballot’’ and uses an LLM to aggregate these ballots.  
Two complementary modes are employed:  
(i) \emph{Independent voting} in which the LLM inspects one Evaluation Agent’s summary table at a time and nominates the best-performing De-id model based on that single table; the final winner is determined by majority across all tables.  
(ii) \emph{Cross-informed voting} in which the LLM simultaneously reviews all Evaluation Agents’ tables and chooses a single best model after considering the combined evidence.  

This dual-mode design serves several purposes.  
First, independent voting captures diverse evaluator perspectives and guards against a single evaluator dominating the decision.  
Second, cross-informed voting enables the LLM to reason globally across evaluators, potentially identifying consensus patterns or compensating for individual evaluator noise.  
By applying natural-language reasoning to structured numerical evidence, LLM majority voting provides an automated and explainable way to integrate heterogeneous assessments.  
It does not re-evaluate the clinical content itself but instead transforms multiple correctness counts into a reliable and reproducible model ranking.

Together, these design elements enable the framework to automatically evaluate and select the best-performing De-id models, even when gold-standard annotations are masked, while maintaining transparency and reproducibility in the aggregation process.

\section{Experimental Settings}

Our experiments evaluate the proposed multi-agent framework for PHI de-identification in clinical notes.  
We focus on two questions: (1) can Evaluation Agents provide consistent and reliable judgments of de-identification quality when gold-standard annotations are hidden, and (2) how well do these judgments align with supervised evaluation and human assessment.

\subsection{Dataset}
We use a dataset of 100 fully annotated clinical notes provided by a large U.S. hospital.  
Each note averages about 1,000 tokens and contains a diverse set of PHI mentions that were meticulously annotated by medical experts.  
While gold-standard labels are available, they are masked in the main experiments and used only for a final supervised check (Table~\ref{tab:gt}).

\subsection{LLM Choices}
Eight LLMs act as De-id models, including commercial APIs and open-source models, as well as two hybrid LPPA models trained on synthetic clinical notes \citep{wu2025large}.  
Six LLMs are employed as Evaluation Agents.  
To ensure fair comparison, all De-id models receive the same prompt and must output a structured JSON dictionary of predicted PHI entities, and Evaluation Agents judge these outputs and return the number of correct predictions in a fixed JSON format.

For readability in our results tables, we use short model names such as \emph{Mistral-7b}, \emph{GPT-3.5}, \emph{GPT-4o}, \emph{Llama-8b}, and \emph{Llama-70b}.


More implementation details—including dataset details, model links, evaluation metrics, validation details, and compute resources—are provided in the Appendix \ref{app:setting}.

\section{Experimental Results}

\subsection{\modelname\ Evaluation Results}
\label{sec:result1}

\paragraph{Single Agent Evaluation.}
Tables \ref{tab:eva_gemma2} to \ref{tab:eva_llama70b}  report results when each of the six Evaluation Agents serves as the judge. Although absolute scores vary considerably across Evaluation Agents, several consistent patterns emerge. Llama-70B is repeatedly judged as one of the top-performing De-id models, appearing in the top-2 in 2 of 6 evaluations (\(\approx\)33\%) and in the top-3 in 4 of 6 evaluations (\(\approx\)67\%).  
It maintains a high average Recall-Proxy (\(\approx\)0.62), strong \textit{Num Correct} values (mean\(\approx\)1,018), and consistently achieves $\geq$0.80 in \textsc{PERSON} recognition, underscoring its stable, evaluator-independent performance. GPT-4o also demonstrates strong performance, particularly in recall-proxy and \textsc{DATE/TIME} recognition. In contrast, smaller models such as GPT-3.5 and Gemma-2 show greater variability: GPT-3.5’s Recall-Proxy spans 0.27–0.87 (mean\(\approx\)0.66, std\(\approx\)0.19), while Gemma-2 spans 0.26–0.69 (mean\(\approx\)0.53, std\(\approx\)0.13). Hybrid models tend to achieve higher precision but exhibit lower coverage, suggesting a trade-off between accuracy and completeness.

\paragraph{Entity-Level Performance.} 
The entity-specific metrics in Tables \ref{tab:eva_gemma2} to \ref{tab:eva_llama70b} highlight additional insights. For the \textsc{PERSON} category, several models, including Llama-70B and GPT-4o, consistently achieve high recognition rates, often exceeding 0.90. This indicates that models are well attuned to detecting personal identifiers such as names. By contrast, performance on \textsc{DATE/TIME} varies more substantially. Some Evaluation Agents judge GPT-4o and Llama-8B as strong performers, while others favor Gemma-2 or Mistral-7B. This divergence reflects differences in how Evaluation Agents interpret temporal expressions, suggesting that date and time recognition remains a challenging subtask for LLM-based de-identification.

\begin{table}[h!]
\centering
\caption{Evaluation Results (Evaluation Agent: Gemma-2)}
\label{tab:eva_gemma2}
\resizebox{0.7\columnwidth}{!}{%
\begin{tabular}{lcccc|cc}
\hline
De-id Model & Precision & Coverage & Num Correct & Recall-Proxy & PERSON & DATE/TIME \\
\hline
Gemma-2      & \cellcolor{red!30}0.6169 & 1.37\% & 984  & 0.5989 & \cellcolor{red!30}0.9120 & \cellcolor{red!30}0.6903 \\
Mistral-7b   & 0.5364 & 1.44\% & 900  & 0.5477 & 0.8626 & \cellcolor{blue!30}0.6754 \\
GPT-3.5       & 0.5513 & \cellcolor{red!30}1.84\% & \cellcolor{red!30}1187 & \cellcolor{red!30}0.7224 & 0.8601 & 0.5928 \\
GPT-4o       & 0.5836 & 1.45\% & 991  & 0.6031 & 0.8659 & 0.5248 \\
Llama-8b     & 0.5555 & \cellcolor{blue!30}1.72\% & \cellcolor{blue!30}1116 & \cellcolor{blue!30}0.6792 & 0.9021 & 0.5975 \\
Llama-70b    & \cellcolor{blue!30}0.5906 & 1.47\% & 1010 & 0.6147 & \cellcolor{blue!30}0.9071 & 0.5981 \\
LPPA4k      & 0.5609 & 1.02\% & 668  & 0.4065 & 0.8778 & 0.5016 \\
LPPA5k      & 0.5743 & 0.95\% & 638  & 0.3883 & 0.8803 & 0.5919 \\
\hline
\end{tabular}%
}
\end{table}

\begin{table}[h!]
\centering
\caption{Evaluation Results (Evaluation Agent: Mistral-7b)}
\resizebox{0.7\columnwidth}{!}{%
\begin{tabular}{lcccc|cc}
\hline
De-id Model & Precision & Coverage & Num Correct & Recall-Proxy & PERSON & DATE/TIME \\
\hline
Gemma-2      & 0.5241 & 1.37\% & 836  & 0.5088 & 0.8944 & 0.4230 \\
Mistral-7b   & 0.5459 & 1.44\% & 916  & 0.5575 & 0.9010 & \cellcolor{red!30}0.6421 \\
GPT-3.5       & 0.4663 & \cellcolor{red!30}1.84\% & \cellcolor{red!30}1004 & \cellcolor{red!30}0.6110 & 0.9336 & 0.4362 \\
GPT-4o       & 0.5595 & 1.45\% & 950  & 0.5782 & 0.9534 & 0.4575 \\
Llama-8b     & 0.4858 & \cellcolor{blue!30}1.72\% & \cellcolor{blue!30}976  & \cellcolor{blue!30}0.5940 & 0.8932 & 0.3363 \\
Llama-70b    & 0.5497 & 1.47\% & 940  & 0.5721 & 0.9214 & 0.5395 \\
LPPA4k      & \cellcolor{blue!30}0.6188 & 1.02\% & 737  & 0.4485 & \cellcolor{red!30}0.9704 & 0.5064 \\
LPPA5k      & \cellcolor{red!30}0.6373 & 0.95\% & 708  & 0.4309 & \cellcolor{blue!30}0.9691 & \cellcolor{blue!30}0.5414 \\
\hline
\end{tabular}%
}
\end{table}

\begin{table}[h!]
\centering
\caption{Evaluation Results (Evaluation Agent: GPT-3.5)}
\resizebox{0.7\columnwidth}{!}{%
\begin{tabular}{lcccc|cc}
\hline
De-id Model & Precision & Coverage & Num Correct & Recall-Proxy & PERSON & DATE/TIME \\
\hline
Gemma-2      & \cellcolor{red!30}0.2742  & 0.91\%  & 431 & 0.2600 & 0.7425 & 0.5609 \\
Mistral-7b   & 0.2301 & 0.97\% & 387 & 0.2335 & 0.7348 & 0.6465 \\
GPT-3.5       & 0.2035 & \cellcolor{red!30}1.26\% & 442 & 0.2667 & 0.8127 & 0.4511 \\
GPT-4o       & 0.2566 & 1.01\% & 446 & 0.2691 & 0.7695 & 0.4470 \\
Llama-8b     & 0.2271 & \cellcolor{blue!30}1.18\% & \cellcolor{blue!30}464 & \cellcolor{blue!30}0.2800 & 0.7681 & 0.5410 \\
Llama-70b    & \cellcolor{blue!30}0.2677 & 1.01\% & \cellcolor{red!30}468 & \cellcolor{red!30}0.2824 & \cellcolor{red!30}0.8358 & 0.4564 \\
LPPA4k      & 0.2620 & 0.69\% & 311 & 0.1876 & \cellcolor{blue!30}0.8282 & \cellcolor{blue!30}0.6646 \\
LPPA5k      & 0.2650 & 0.65\% & 296 & 0.1786 & 0.8221 & \cellcolor{red!30}0.6903 \\
\hline
\end{tabular}%
}
\end{table}

\begin{table}[h!]
\centering
\caption{Evaluation Results (Evaluation Agent: GPT-4o)}
\resizebox{0.7\columnwidth}{!}{%
\begin{tabular}{lcccc|cc}
\hline
De-id Model & Precision & Coverage & Num Correct & Recall-Proxy & PERSON & DATE/TIME \\
\hline
Gemma-2      & 0.6177 & 0.91\% & 971 & 0.5859 & 0.6343 & 0.9013 \\
Mistral-7b   & 0.5279 & 0.97\% & 888 & 0.5358 & 0.6796 & 0.9130 \\
GPT-3.5      & 0.5787 & \cellcolor{red!30}1.26\% & \cellcolor{red!30}1257 & \cellcolor{red!30}0.7584 & 0.6890 & 0.8324 \\
GPT-4o       & 0.6559 & 1.01\% & 1140 & 0.6878 & 0.6796 & 0.8874 \\
Llama-8b     & 0.6089 & \cellcolor{blue!30}1.18\% & \cellcolor{blue!30}1244 & \cellcolor{blue!30}0.7506 & 0.6777 & 0.9306 \\
Llama-70b    & \cellcolor{red!30}0.7094 & 1.01\% & 1240 & 0.7482 & \cellcolor{red!30}0.8139 & 0.9071 \\
LPPA4k      & 0.6664 & 0.69\% & 791 & 0.4773 & 0.7786 & \cellcolor{red!30}0.9541 \\
LPPA5k      & \cellcolor{blue!30}0.6723 & 0.65\% & 751 & 0.4531 & \cellcolor{blue!30}0.7905 & \cellcolor{blue!30}0.9428 \\
\hline
\end{tabular}%
}
\end{table}

\begin{table}[h!]
\centering
\caption{Evaluation Results (Evaluation Agent: Llama-8b)}
\resizebox{0.7\columnwidth}{!}{%
\begin{tabular}{lcccc|cc}
\hline
De-id Model & Precision & Coverage & Num Correct & Recall-Proxy & PERSON & DATE/TIME \\
\hline
Gemma-2      & 0.5808 & 0.91\% & 913 & 0.5509 & 0.9552 & \cellcolor{red!30}0.7731 \\
Mistral-7b   & 0.5511 & 0.97\% & 927 & 0.5593 & 0.9724 & 0.6554 \\
GPT-3.5       & 0.5506 & \cellcolor{red!30}1.26\% & \cellcolor{red!30}1196 & \cellcolor{red!30}0.7216 & 0.9223 & 0.6940 \\
GPT-4o       & 0.6007 & 1.01\% & 1044 & 0.6299 & 0.9581 & 0.6800 \\
Llama-8b     & 0.5546 & \cellcolor{blue!30}1.18\% & \cellcolor{blue!30}1133 & \cellcolor{blue!30}0.6836 & 0.9699 & 0.7087 \\
Llama-70b    & 0.5749 & 1.01\% & 1005 & 0.6064 & \cellcolor{blue!30}0.9745 & 0.6750 \\
LPPA4k      & \cellcolor{blue!30}0.6318 & 0.69\% & 750 & 0.4525 & \cellcolor{red!30}0.9771 & \cellcolor{blue!30}0.7468 \\
LPPA5k      & \cellcolor{red!30}0.6741 & 0.65\% & 753 & 0.4543 & 0.9684 & 0.6627 \\
\hline
\end{tabular}%
}
\end{table}

\begin{table}[h!]
\centering
\caption{Evaluation Results (Evaluation Agent: Llama-70b)}
\label{tab:eva_llama70b}
\resizebox{0.7\columnwidth}{!}{%
\begin{tabular}{lcccc|cc}
\hline
De-id Model & Precision & Coverage & Num Correct & Recall-Proxy & PERSON & DATE/TIME \\
\hline
Gemma-2      & 0.7284 & 0.91\% & 1145 & 0.6909 & 0.9478 & 0.9118 \\
Mistral-7b   & 0.6599 & 0.97\% & 1110 & 0.6697 & 0.8564 & 0.8845 \\
GPT-3.5       & 0.6653 & \cellcolor{red!30}1.26\% & \cellcolor{blue!30}1445 & 0.8719 & 0.9117 & 0.8571 \\
GPT-4o       & 0.8096 & 1.01\% & 1407 & 0.8489 & 0.9551 & 0.8852 \\
Llama-8b     & 0.7347 & \cellcolor{blue!30}1.18\% & \cellcolor{red!30}1501 & \cellcolor{red!30}0.9056 & \cellcolor{blue!30}0.9578 & 0.8941 \\
Llama-70b    & \cellcolor{red!30}0.8278 & 1.01\% & 1447 & \cellcolor{blue!30}0.8731 & \cellcolor{red!30}1.0000 & 0.9241 \\
LPPA4k      & 0.7944 & 0.69\% & 943 & 0.5690 & 0.9313 & \cellcolor{blue!30}0.9241 \\
LPPA5k      & \cellcolor{blue!30}0.8147 & 0.65\% & 910 & 0.5491 & 0.9447 & \cellcolor{red!30}0.9566 \\
\hline
\end{tabular}%
}
\end{table}

\paragraph{LLM Majority Voting.}

\begin{table}[htbp]
\centering
\scriptsize
\caption{LLM voting: best De-id model selected independently or with cross-informed context.}
\label{tab:llm_vote}
\resizebox{0.5\columnwidth}{!}{
\begin{tabular}{lcc}
\toprule
Models & Independent & Cross-Informed \\
\midrule
Gemma-2    & Llama-70b & Llama-70b \\
Mistral-7b & Llama-70b & Llama-70b \\
GPT-3.5    & Llama-70b & Llama-70b \\
GPT-4o     & Llama-70b & Llama-70b \\
Llama-8b   & Llama-8b  & Llama-70b \\
Llama-70b  & Llama-8b  & Llama-70b \\
\bottomrule
\end{tabular}
}
\end{table}

To reduce individual evaluator bias of De-id models, we apply the LLM majority voting procedure described in Section~\ref{sec:multi-agent-evaluation}.  
This procedure aggregates the numerical judgments produced by multiple Evaluation Agents and asks an LLM to determine the best-performing De-id model under two complementary modes:
(i) \emph{Independent voting}, in which the voting LLM reviews one Evaluation Agent’s summary table at a time and casts a vote for the best model on that table; and
(ii) \emph{Cross-informed voting}, in which the voting LLM reviews all Evaluation Agents’ tables together and selects the best model after considering the combined evidence.
The two modes capture different perspectives: independent voting prevents a single evaluator from dominating the decision, while cross-informed voting allows global reasoning across evaluators.  

As shown in Table~\ref{tab:llm_vote}, both voting schemes consistently select Llama-70B as the top-performing De-id model. Under independent voting, 4 of 6 evaluation agents (67\%) selected Llama-70B as best, while under cross-informed voting, all 6 of 6 agents (100\%) agreed on Llama-70B. This high level of agreement indicates that ensemble judgments are more stable and less biased than those derived from any single evaluator and that LLM majority voting provides a robust mechanism to automatically evaluate and select the best De-id models when gold-standard annotations are unavailable.

\subsection{External Validation}
\label{sec:result2}

\begin{table}[htbp]
\centering
\caption{Ground Truth evaluation results.}
\renewcommand{\arraystretch}{1.8} 
\resizebox{1\columnwidth}{!}{%
\begin{tabular}{lccccccccc}
\hline
\textbf{Models} & \multicolumn{3}{c}{Overall} & \multicolumn{3}{c}{PERSON} & \multicolumn{3}{c}{DATE/TIME} \\
\cline{2-10}
 & Pr & Re & F1 & Pr & Re & F1 & Pr & Re & F1 \\
\hline
Gemma-2              & 0.52$\pm$0.01                                 & 0.45$\pm$0.01                                 & 0.48$\pm$0.01                                 & 0.53$\pm$0.01                                & 0.47$\pm$0.01                             & 0.50$\pm$0.02                                 & \cellcolor{red!30}0.88$\pm$0.02                             & 0.41$\pm$0.01                                 & 0.56$\pm$0.02                                 \\
Mistral-7b          & 0.52$\pm$0.01                                 & 0.50$\pm$0.01                                 & 0.51$\pm$0.01                                 & 0.58$\pm$0.01                                & 0.36$\pm$0.01                             & 0.44$\pm$0.02                                 & \cellcolor{blue!30}0.87$\pm$0.02                             & 0.50$\pm$0.01                                 & 0.63$\pm$0.02                                 \\
GPT-3.5       & 0.43$\pm$0.02 & 0.60$\pm$0.01 & 0.48$\pm$0.02 & \cellcolor{blue!30}0.60$\pm$0.04 & 0.50$\pm$0.01 & 0.54$\pm$0.02 & 0.74$\pm$0.03 & 0.44$\pm$0.02\textsuperscript{*} & 0.55$\pm$0.02 \\
GPT-4o               & 0.53$\pm$0.01$^*$ & \cellcolor{red!30}0.69$^*$ & \cellcolor{blue!30}0.58$\pm$0.01$^*$ & \cellcolor{red!30}0.60$\pm$0.02$^*$ & \cellcolor{red!30}0.57$\pm$0.01 & \cellcolor{red!30}0.58$\pm$0.01$^*$ & 0.80$\pm$0.01 & \cellcolor{red!30}0.57$\pm$0.02$^*$ & \cellcolor{red!30}0.67$\pm$0.02$^*$ \\
Llama-8b  & 0.46$\pm$0.02 & 0.59$\pm$0.03 & 0.50$\pm$0.02 & 0.53$\pm$0.01 & \cellcolor{blue!30}0.55$\pm$0.02 & 0.53$\pm$0.01 & 0.79$\pm$0.03 & 0.39$\pm$0.02 & 0.52$\pm$0.01 \\
Llama-70b & 0.60$\pm$0.01$^*$ & \cellcolor{blue!30}0.68$\pm$0.01$^*$ & \cellcolor{red!30}0.62$\pm$0.01$^*$ & 0.59$\pm$0.01$^*$ & 0.53$\pm$0.01 & 0.56$\pm$0.01$^*$ & 0.83$\pm$0.01 & \cellcolor{blue!30}0.50$\pm$0.01$^*$ & \cellcolor{blue!30}0.63$\pm$0.01$^*$ \\
LPPA4k     & \cellcolor{red!30}0.65$\pm$0.01$^*$ & 0.54$\pm$0.02 & 0.57$\pm$0.01$^*$ & 0.59$\pm$0.02$^*$ & 0.53$\pm$0.02 & 0.56$\pm$0.02 & 0.82$\pm$0.03 & 0.40$\pm$0.03 & 0.54$\pm$0.03 \\
LPPA5k     & \cellcolor{blue!30}0.64$\pm$0.01$^*$ & 0.55$\pm$0.02 & 0.57$\pm$0.01$^*$ & 0.59$\pm$0.03$^*$ & 0.53$\pm$0.02 & \cellcolor{blue!30}0.56$\pm$0.01$^*$ & 0.82$\pm$0.02 & 0.35$\pm$0.02 & 0.50$\pm$0.01 \\
\hline
\end{tabular}%
}
\label{tab:gt}
\end{table}

\begin{table}[htbp]
\centering
\scriptsize
\caption{Human evaluation of Evaluation Agents. Q1: correctness; Q2: missed-PHI severity; Q3: overall trustworthiness.}
\label{tab:human_eval}
\resizebox{0.5\columnwidth}{!}{
\begin{tabular}{lccc}
\toprule
Models & Q1 ($\uparrow$) & Q2 ($\downarrow$) & Q3 ($\uparrow$) \\
\midrule
Gemma-2    & 0.00 & 2.20 & 1.60 \\
Mistral-7b & 0.00 & 2.24 & 1.40 \\
GPT-3.5    & 0.00 & 4.28 & 1.68 \\
GPT-4o     & \cellcolor{blue!30}0.18 & 2.22 & \cellcolor{blue!30}3.85 \\
Llama-8b   & \cellcolor{red!30}0.20 & \cellcolor{blue!30}2.08 & 3.75 \\
Llama-70b  & 0.18 & \cellcolor{red!30}1.62 & \cellcolor{red!30}4.20 \\
\bottomrule
\end{tabular}
}
\end{table}

\paragraph{Ground-truth Evaluation.}
Although the multi-agent framework is designed to operate without human-labeled data, we also perform \emph{ground-truth evaluation} to further validate the evaluation conclusions of \modelname\ described in Section \ref{sec:result1}.  
Table~\ref{tab:gt} reports precision, recall, and F1 scores computed against manually annotated PHI labels.  
Llama-70B achieves the highest overall F1 (0.62), followed by GPT-4o (0.58) and the hybrid models (0.57).  
The agreement between ground-truth evaluation and LLM-based judging is striking: models ranked highest by Evaluation Agents and LLM majority voting are the same models that perform best under gold-standard supervision.  
This demonstrates that the proposed framework can reproduce supervised rankings even when gold annotations are hidden.

\paragraph{Human Evaluation.}
To further verify the quality of Evaluation Agents’ judgments, we also conduct a \emph{human evaluation} of a representative sample of notes.  
Two biomedical experts, supported by LLM-generated checklists and highlights, inspect each case to confirm correctness counts, estimate the number of missed PHI entities, and rate overall trustworthiness.  
As summarized in Table~\ref{tab:human_eval}, human reviewers consistently rate Llama-70B outputs highest on overall quality.  
It achieves the lowest missed-PHI severity (Q2 = 1.62, best among all models) and the highest overall trustworthiness (Q3 = 4.20), while tying with GPT-4o for second place in correctness (Q1 = 0.18; Llama-8B leads slightly at 0.20).  
All other models have Q1 $\le$ 0.18 and Q3 $\le$ 3.85, underscoring the clear human preference for Llama-70B.

Together, these supervised and human assessments demonstrate that our multi-agent framework can reliably approximate gold-standard evaluation and produce reproducible, trustworthy rankings of De-id models even when ground-truth annotations are intentionally withheld.

\subsection{Summary of Findings.} 
Overall, the experiments demonstrate that while Evaluation Agents differ in absolute scoring, they consistently reveal relative performance trends. Llama-70B emerges as the most reliable De-id model. Hybrid models offer precision advantages but at the cost of coverage. Entity-specific analysis shows that \textsc{PERSON} recognition is relatively robust, while \textsc{DATE/TIME} detection remains less stable. The convergence of LLM-based evaluation, gold-standard comparison, voting, and human judgment underscores the robustness of our proposed framework and highlights its practical utility in real-world clinical text processing where annotated datasets are unavailable.

\section{Conclusion}

We introduced \modelname, a multi-agent framework that enables automatic evaluation and selection of PHI de-identification models in clinical notes without heavy reliance on manually annotated data.  
In this framework, large language models (LLMs) operate as independent \textit{Evaluation Agents} that assess de-identification quality, while their judgments are consolidated through LLM-based majority voting to produce stable and reproducible model rankings.

Experiments on a real-world clinical-note corpus demonstrate that \modelname\ provides consistent and trustworthy evaluations: despite variation across individual evaluators, majority voting converges on the same top-performing systems, with Llama-70B and GPT-4o repeatedly identified as leading models.  
Entity-level analysis further shows that recognition of \textsc{PERSON} entities is highly stable, whereas \textsc{DATE/TIME} remains more variable, suggesting areas for targeted improvement.

Validation against a manually annotated test set confirms that the rankings produced by \modelname\ closely align with ground-truth evaluation, and additional human review reinforces the reliability of the automated judgments.  

Overall, this work establishes that LLMs can act not only as PHI de-identifiers but also as scalable and dependable evaluators.  
\modelname\ provides three key contributions: (1) a practical framework for benchmarking and selecting PHI de-identification models without extensive human labeling; (2) empirical evidence that multi-agent evaluation and majority voting effectively mitigate individual evaluator bias; and (3) fine-grained insights that guide refinement of PHI coverage and accuracy.  
Future work will extend this framework to additional PHI categories and larger datasets, and explore techniques to better calibrate Evaluation Agents and integrate limited human oversight for even greater robustness.

\newpage

\section*{AI Agent Setup}

In this work, we used OpenAI's GPT-5 model as the sole AI agent to support experimental design, implementation, data analysis, and manuscript preparation. All interactions with GPT-5 were conducted through the official web interface, without employing orchestration frameworks such as LangChain or AutoGen. The model generated initial implementations of the multi-agent evaluation framework, suggested experimental setups, analyzed results, and drafted sections of the paper. Human authors executed and validated the AI-generated code, refined analyses, and ensured the accuracy of interpretations. No external tools were integrated beyond standard Python environments and model inference interfaces described in the Experimental Settings. All de-identification and evaluation experiments were conducted using the LLMs listed in Appendix~\ref{app:setting}, through a combination of API-based and local runs on the Azure platform. This streamlined configuration shows that a single, general-purpose LLM agent can effectively support the entire research workflow for PHI de-identification evaluation without complex orchestration or additional infrastructure.


\bibliographystyle{plainnat}   
\bibliography{references}      

@article{wu2025large,
  title={Large Language Model Empowered Privacy-Protected Framework for PHI Annotation in Clinical Notes},
  author={Wu, Guanchen and Zheng, Linzhi and Xie, Han and Xiang, Zhen and Lu, Jiaying and Liu, Darren and Bold, Delgersuren and Li, Bo and Hu, Xiao and Yang, Carl},
  journal={arXiv preprint arXiv:2504.18569},
  year={2025}
}

@article{kovavcevic2024identification,
  title={De-identification of clinical free text using natural language processing: A systematic review of current approaches},
  author={Kova{\v{c}}evi{\'c}, Aleksandar and Ba{\v{s}}aragin, Bojana and Milo{\v{s}}evi{\'c}, Nikola and Nenadi{\'c}, Goran},
  journal={Artificial intelligence in medicine},
  volume={151},
  pages={102845},
  year={2024},
  publisher={Elsevier}
}

@article{uzuner2007evaluating,
  title={Evaluating the state-of-the-art in automatic de-identification},
  author={Uzuner, {\"O}zlem and Luo, Yuan and Szolovits, Peter},
  journal={Journal of the American Medical Informatics Association},
  volume={14},
  number={5},
  pages={550--563},
  year={2007},
  publisher={BMJ Group BMA House, Tavistock Square, London, WC1H 9JR}
}

@article{meystre2010automatic,
  title={Automatic de-identification of textual documents in the electronic health record: a review of recent research},
  author={Meystre, Stephane M and Friedlin, F Jeffrey and South, Brett R and Shen, Shuying and Samore, Matthew H},
  journal={BMC medical research methodology},
  volume={10},
  number={1},
  pages={70},
  year={2010},
  publisher={Springer}
}

@inproceedings{zhang2024tacco,
  title={Tacco: Task-guided co-clustering of clinical concepts and patient visits for disease subtyping based on ehr data},
  author={Zhang, Ziyang and Cui, Hejie and Xu, Ran and Xie, Yuzhang and Ho, Joyce C and Yang, Carl},
  booktitle={Proceedings of the 30th ACM SIGKDD Conference on Knowledge Discovery and Data Mining},
  pages={6324--6334},
  year={2024}
}

@article{stubbs2015automated,
  title={Automated systems for the de-identification of longitudinal clinical narratives: Overview of 2014 i2b2/UTHealth shared task Track 1},
  author={Stubbs, Amber and Kotfila, Christopher and Uzuner, {\"O}zlem},
  journal={Journal of biomedical informatics},
  volume={58},
  pages={S11--S19},
  year={2015},
  publisher={Elsevier}
}

@article{he2015crfs,
  title={CRFs based de-identification of medical records},
  author={He, Bin and Guan, Yi and Cheng, Jianyi and Cen, Keting and Hua, Wenlan},
  journal={Journal of biomedical informatics},
  volume={58},
  pages={S39--S46},
  year={2015},
  publisher={Elsevier}
}

@article{jiang2017identification,
  title={De-identification of medical records using conditional random fields and long short-term memory networks},
  author={Jiang, Zhipeng and Zhao, Chao and He, Bin and Guan, Yi and Jiang, Jingchi},
  journal={Journal of biomedical informatics},
  volume={75},
  pages={S43--S53},
  year={2017},
  publisher={Elsevier}
}

@article{dernoncourt2017identification,
  title={De-identification of patient notes with recurrent neural networks},
  author={Dernoncourt, Franck and Lee, Ji Young and Uzuner, Ozlem and Szolovits, Peter},
  journal={Journal of the American Medical Informatics Association},
  volume={24},
  number={3},
  pages={596--606},
  year={2017},
  publisher={Oxford University Press}
}

@inproceedings{tang2020identification,
  title={De-identification of clinical text via Bi-LSTM-CRF with neural language models},
  author={Tang, Buzhou and Jiang, Dehuan and Chen, Qingcai and Wang, Xiaolong and Yan, Jun and Shen, Ying},
  booktitle={AMIA Annual Symposium Proceedings},
  volume={2019},
  pages={857},
  year={2020}
}

@inproceedings{johnson2020deidentification,
  title={Deidentification of free-text medical records using pre-trained bidirectional transformers},
  author={Johnson, Alistair EW and Bulgarelli, Lucas and Pollard, Tom J},
  booktitle={Proceedings of the ACM conference on health, inference, and learning},
  pages={214--221},
  year={2020}
}

@article{lee2020biobert,
  title={BioBERT: a pre-trained biomedical language representation model for biomedical text mining},
  author={Lee, Jinhyuk and Yoon, Wonjin and Kim, Sungdong and Kim, Donghyeon and Kim, Sunkyu and So, Chan Ho and Kang, Jaewoo},
  journal={Bioinformatics},
  volume={36},
  number={4},
  pages={1234--1240},
  year={2020},
  publisher={Oxford University Press}
}

@article{liu2023deid,
  title={Deid-gpt: Zero-shot medical text de-identification by gpt-4},
  author={Liu, Zhengliang and Huang, Yue and Yu, Xiaowei and Zhang, Lu and Wu, Zihao and Cao, Chao and Dai, Haixing and Zhao, Lin and Li, Yiwei and Shu, Peng and others},
  journal={arXiv preprint arXiv:2303.11032},
  year={2023}
}

@article{heider2024extensible,
  title={An extensible evaluation framework applied to clinical text deidentification natural language processing tools: multisystem and multicorpus study},
  author={Heider, Paul M and Meystre, St{\'e}phane M},
  journal={Journal of medical Internet research},
  volume={26},
  pages={e55676},
  year={2024},
  publisher={JMIR Publications Toronto, Canada}
}

@inproceedings{xie2024promptlink,
  title={PromptLink: leveraging large language models for cross-source biomedical concept linking},
  author={Xie, Yuzhang and Lu, Jiaying and Ho, Joyce and Nahab, Fadi and Hu, Xiao and Yang, Carl},
  booktitle={Proceedings of the 47th International ACM SIGIR Conference on Research and Development in Information Retrieval},
  pages={2589--2593},
  year={2024}
}

@article{bhasuran2025preliminary,
  title={Preliminary analysis of the impact of lab results on large language model generated differential diagnoses},
  author={Bhasuran, Balu and Jin, Qiao and Xie, Yuzhang and Yang, Carl and Hanna, Karim and Costa, Jennifer and Shavor, Cindy and Han, Wenshan and Lu, Zhiyong and He, Zhe},
  journal={npj Digital Medicine},
  volume={8},
  number={1},
  pages={166},
  year={2025},
  publisher={Nature Publishing Group UK London}
}

@article{zhu2023judgelm,
  title={Judgelm: Fine-tuned large language models are scalable judges},
  author={Zhu, Lianghui and Wang, Xinggang and Wang, Xinlong},
  journal={arXiv preprint arXiv:2310.17631},
  year={2023}
}

@inproceedings{kim2023prometheus,
  title={Prometheus: Inducing fine-grained evaluation capability in language models},
  author={Kim, Seungone and Shin, Jamin and Cho, Yejin and Jang, Joel and Longpre, Shayne and Lee, Hwaran and Yun, Sangdoo and Shin, Seongjin and Kim, Sungdong and Thorne, James and others},
  booktitle={The Twelfth International Conference on Learning Representations},
  year={2023}
}

@article{xie2025kerap,
  title={KERAP: A Knowledge-Enhanced Reasoning Approach for Accurate Zero-shot Diagnosis Prediction Using Multi-agent LLMs},
  author={Xie, Yuzhang and Cui, Hejie and Zhang, Ziyang and Lu, Jiaying and Shu, Kai and Nahab, Fadi and Hu, Xiao and Yang, Carl},
  journal={arXiv preprint arXiv:2507.02773},
  year={2025}
}

@article{hada2023large,
  title={Are large language model-based evaluators the solution to scaling up multilingual evaluation?},
  author={Hada, Rishav and Gumma, Varun and de Wynter, Adrian and Diddee, Harshita and Ahmed, Mohamed and Choudhury, Monojit and Bali, Kalika and Sitaram, Sunayana},
  journal={arXiv preprint arXiv:2309.07462},
  year={2023}
}

@article{wang2023large,
  title={Large language models are not fair evaluators},
  author={Wang, Peiyi and Li, Lei and Chen, Liang and Cai, Zefan and Zhu, Dawei and Lin, Binghuai and Cao, Yunbo and Liu, Qi and Liu, Tianyu and Sui, Zhifang},
  journal={arXiv preprint arXiv:2305.17926},
  year={2023}
}

@article{seinen2025using,
  title={Using structured codes and free-text notes to measure information complementarity in electronic health records: Feasibility and validation study},
  author={Seinen, Tom M and Kors, Jan A and van Mulligen, Erik M and Rijnbeek, Peter R},
  journal={Journal of Medical Internet Research},
  volume={27},
  pages={e66910},
  year={2025},
  publisher={JMIR Publications Toronto, Canada}
}

@article{tayefi2021challenges,
  title={Challenges and opportunities beyond structured data in analysis of electronic health records},
  author={Tayefi, Maryam and Ngo, Phuong and Chomutare, Taridzo and Dalianis, Hercules and Salvi, Elisa and Budrionis, Andrius and Godtliebsen, Fred},
  journal={Wiley Interdisciplinary Reviews: Computational Statistics},
  volume={13},
  number={6},
  pages={e1549},
  year={2021},
  publisher={Wiley Online Library}
}

@article{cohen2018hipaa,
  title={HIPAA and protecting health information in the 21st century},
  author={Cohen, I Glenn and Mello, Michelle M},
  journal={Jama},
  volume={320},
  number={3},
  pages={231--232},
  year={2018},
  publisher={American Medical Association}
}

@article{moore2019review,
  title={Review of HIPAA, part 1: history, protected health information, and privacy and security rules},
  author={Moore, Wilnellys and Frye, Sarah},
  journal={Journal of nuclear medicine technology},
  volume={47},
  number={4},
  pages={269--272},
  year={2019},
  publisher={Society of Nuclear Medicine}
}

@inproceedings{lehman2023we,
  title={Do we still need clinical language models?},
  author={Lehman, Eric and Hernandez, Evan and Mahajan, Diwakar and Wulff, Jonas and Smith, Micah J and Ziegler, Zachary and Nadler, Daniel and Szolovits, Peter and Johnson, Alistair and Alsentzer, Emily},
  booktitle={Conference on health, inference, and learning},
  pages={578--597},
  year={2023},
  organization={PMLR}
}

@article{wu2024ontology,
  title={Ontology extension by online clustering with large language model agents},
  author={Wu, Guanchen and Ling, Chen and Graetz, Ilana and Zhao, Liang},
  journal={Frontiers in Big Data},
  volume={7},
  pages={1463543},
  year={2024},
  publisher={Frontiers Media SA}
}

@article{altalla2025evaluating,
  title={Evaluating GPT models for clinical note de-identification},
  author={Altalla’, Bayan and Abdalla, Sameera and Altamimi, Ahmad and Bitar, Layla and Al Omari, Amal and Kardan, Ramiz and Sultan, Iyad},
  journal={Scientific Reports},
  volume={15},
  number={1},
  pages={3852},
  year={2025},
  publisher={Nature Publishing Group UK London}
}

@article{pan2024graphnarrator,
  title={GraphNarrator: Generating Textual Explanations for Graph Neural Networks},
  author={Pan, Bo and Xiong, Zhen and Wu, Guanchen and Zhang, Zheng and Zhang, Yifei and Zhao, Liang},
  journal={arXiv preprint arXiv:2410.15268},
  year={2024}
}

@article{de2025study,
  title={A study of calibration as a measurement of trustworthiness of large language models in biomedical natural language processing},
  author={de Oliveira, Rodrigo and Garber, Matthew and Gwinnutt, James M and Rashidi, Emaan and Hwang, Jwu-Hsuan and Gilmour, William and Nanavati, Jay and Zine El Abidine, Khaldoun and Mack, Christina DeFilippo},
  journal={JAMIA open},
  volume={8},
  number={4},
  pages={ooaf058},
  year={2025},
  publisher={Oxford University Press}
}

@article{chang2024survey,
  title={A survey on evaluation of large language models},
  author={Chang, Yupeng and Wang, Xu and Wang, Jindong and Wu, Yuan and Yang, Linyi and Zhu, Kaijie and Chen, Hao and Yi, Xiaoyuan and Wang, Cunxiang and Wang, Yidong and others},
  journal={ACM transactions on intelligent systems and technology},
  volume={15},
  number={3},
  pages={1--45},
  year={2024},
  publisher={ACM New York, NY}
}

\appendix
\newpage
\section*{Appendix}

\section{Notation Table}
\label{app:notation}
The notations in this paper are summarized in Table~\ref{tab:notation}.
\begin{table}[htbp]
\centering
\caption{Notation used in the paper.}
\label{tab:notation}
\begin{tabular}{@{}ll@{}}
\toprule
Symbol & Description \\ 
\midrule
$\mathcal{X}$ & Space of clinical notes. \\
$\mathcal{C}=\{c_1,\ldots,c_m\}$ & Set of PHI categories; $m$ is the number of categories. \\
$x\in\mathcal{X}$ & A clinical note. \\
$c_i\in\mathcal{C}$ & A PHI category label. \\
$e_i\subseteq x$ & A contiguous text span (entity) in $x$. \\
$k$ & Number of predicted entities in $\texttt{PHI}(x)$. \\
$\texttt{PHI}(x)$ & Set of predicted (category, span) pairs for $x$, $\{(c_i,e_i)\}_{i=1}^k$. \\
$\mathcal{M}=\{M_1,\ldots,M_R\}$ & Candidate set of de-identification models; $R$ is its size. \\
$M,\,M^*$ & A candidate model; $M^*$ is the selected (best) model by Eq.~\eqref{eqa:select}. \\
$\mathcal{D}$ & Data distribution over notes used in the expectation in Eq.~\eqref{eqa:select}. \\
$\texttt{PHI\mbox{-}EV}(M,x)$ & Evaluation score for model $M$ on input $x$. \\
$\texttt{PHI}_M(x)$ & Prediction of model $M$ on $x$ (also written as $\texttt{PHI}(M,x)$). \\
$y^*(x)$ & Gold-standard annotation for $x$ (used in conventional evaluation). \\
$f$ & Supervised metric comparing $\texttt{PHI}_M(x)$ and $y^*(x)$ (e.g., precision/recall/F1). \\
$A$ & Number of LLM-based Evaluation Agents. \\
$E_a$ & The $a$-th Evaluation Agent, $a=1,\ldots,A$. \\
$g$ & Multi-agent evaluator that aggregates $\{E_a\}_{a=1}^A$ (e.g., majority voting). \\
$\mathrm{Aggregate}(\cdot)$ & Aggregation operator combining agent judgments. \\
\bottomrule
\end{tabular}
\end{table}





\section{Additional Experimental Details}
\label{app:setting}

\subsection{Dataset Details}
Access to real-world clinical notes is highly restricted by privacy regulations, and fully annotated datasets are expensive and rare.  
For this study we obtained 100 authentic clinical notes, each averaging about 1{,}000 tokens, in which all PHI entities were meticulously annotated by multiple medical experts.  
These notes retain their original structure and contain diverse, detailed patient information.  
Although gold-standard annotations exist, we masked them in the main experiments to test whether our proposed framework can assess PHI de-identification performance without relying on annotated references.  
The gold labels were used only for a separate supervised evaluation (Table~\ref{tab:gt}) to externally validate findings derived from LLM-based judgments.

\subsection{Model Links}
For reproducibility, we provide the URLs of all language models used as De-id models or Evaluation Agents:
\begin{itemize}
    \item Gemma-2: \href{https://huggingface.co/google/gemma-2-9b-it}{https://huggingface.co/google/gemma-2-9b-it}
    \item Mistral-7B-Instruct: \href{https://huggingface.co/mistralai/Mistral-7B-Instruct-v0.3}{https://huggingface.co/mistralai/Mistral-7B-Instruct-v0.3}
    \item Llama-3-8B-Instruct: \href{https://huggingface.co/meta-llama/Meta-Llama-3-8B-Instruct}{https://huggingface.co/meta-llama/Meta-Llama-3-8B-Instruct}
    \item Llama-3-70B-Instruct: \href{https://huggingface.co/meta-llama/Meta-Llama-3-70B-Instruct}{https://huggingface.co/meta-llama/Meta-Llama-3-70B-Instruct}
    \item GPT-3.5-turbo-0125: \href{https://platform.openai.com/docs/models/gpt-3.5-turbo}{https://platform.openai.com/docs/models/gpt-3.5-turbo}
    \item GPT-4o-mini: \href{https://platform.openai.com/docs/models/gpt-4o-mini}{https://platform.openai.com/docs/models/gpt-4o-mini}
    \item LPPA4k: \href{https://huggingface.co/spacebetweenus/108mix4ktest1}{https://huggingface.co/spacebetweenus/108mix4ktest1}
    \item LPPA5k: \href{https://huggingface.co/spacebetweenus/107mix5k}{https://huggingface.co/spacebetweenus/107mix5k}
\end{itemize}

\subsection{Evaluation Metrics}
We report four core metrics to quantify De-id models performance under Evaluation Agents:
\begin{align*}
\text{Precision}_d &= \frac{|C_d|}{|P_d|}, \\
\text{Coverage}_d &= \frac{\sum_n |P_d(n)|}{\sum_n T(n)}, \\
\text{CorrectCount}_d &= |C_d|, \\
\text{RecallProxy}_d &= \frac{|C_d|}{N_{\text{avg}}}.
\end{align*}
Here $P_d$ is the set of PHI predictions by De-id model $d$, $C_d$ the subset judged correct, $T(n)$ the token count of note $n$, and $N_{\text{avg}}$ the average number of PHI predictions across all agents.  
We also compute category-specific precision for \textsc{PERSON} and \textsc{DATE/TIME}, and in a separate supervised evaluation (Table~\ref{tab:gt}) compute true precision, recall, and F1 against gold annotations.


\subsection{Validation and Human Assessment}

To confirm the reliability of the multi-agent evaluation, we performed two complementary validations.  
(i) We conducted a direct supervised evaluation of all de-identification models against the gold-standard annotations (Table~\ref{tab:gt}) to verify that the rankings produced by our framework align with conventional precision, recall, and F1 metrics.  
(ii) We carried out a human expert review of a representative subset of Evaluation Agent outputs (Table~\ref{tab:human_eval}), in which biomedical experts assessed correctness, missed-PHI severity, and overall trustworthiness.  

These validations demonstrate that the multi-agent evaluation and LLM majority voting used in our main experiments provide rankings that are consistent with both gold-standard supervision and independent human judgment.

\subsection{Compute Resources}
All experiments with Llama models (Llama-8B and Llama-70B) and LPPA hybrid models (LPPA4k and LPPA5k) were conducted on a server equipped with two NVIDIA H100 GPUs (80GB memory each). Experiments with GPT models (GPT-3.5 and GPT-4o) were performed through the Azure OpenAI Service, which provides HIPAA-compliant secure inference. Smaller open-source models, including Gemma-2 and Mistral-7B, were run locally on an Apple MacBook (M2 Pro, 32GB RAM). These experiments were inference-only; each H100 run required less than two hours and the total compute time was about 200 GPU-hours. API-based GPT experiments incurred standard usage costs but required no additional hardware.

\section{Limitations}
This study has several limitations. First, evaluation was conducted on a single set of 100 annotated clinical notes, which may not capture the diversity of clinical documentation across institutions, specialties, or languages; larger and more heterogeneous datasets are needed to confirm generalizability. Second, our framework assumes that cross-agent agreement is a reliable proxy for recall when ground truth is masked. This may be violated if multiple agents share systematic biases (e.g., under-detecting specific PHI types), potentially inflating recall-proxy values. Third, performance depends on practical factors such as prompt design, model size, and context length; small prompt or tokenization changes can affect results, and while majority voting reduces variance, complete stability is not guaranteed. Fourth, deploying multiple large LLMs for both de-identification and evaluation incurs significant computational cost, which limits scalability and real-time use; lightweight or distilled models could address this. Finally, although designed for privacy protection, the framework does not inherently ensure fairness or mitigate demographic or linguistic bias, and it targets PHI categories defined under U.S. HIPAA regulations, requiring adaptation for other legal or linguistic contexts.








\newpage

\section*{Agents4Science AI Involvement Checklist}

\begin{enumerate}
    \item \textbf{Hypothesis development}: Hypothesis development includes the process by which you came to explore this research topic and research question. This can involve the background research performed by either researchers or by AI. This can also involve whether the idea was proposed by researchers or by AI. 

    Answer: \involvementA{} 
    
    Explanation: The authors proposed the TEAM-PHI framework and formulated the research problem, including multi-agent evaluation and LLM majority voting, through human background research and scientific reasoning. LLMs were not used to generate hypotheses or define research questions. 
    \item \textbf{Experimental design and implementation}: This category includes design of experiments that are used to test the hypotheses, coding and implementation of computational methods, and the execution of these experiments. 

    Answer: \involvementC{} 
    
    Explanation: Large language models designed the experimental framework, generated the prompts, and produced the complete code and implementation details. Human authors’ role was limited to executing the AI-generated code and verifying that it ran successfully. 
    \item \textbf{Analysis of data and interpretation of results}: This category encompasses any process to organize and process data for the experiments in the paper. It also includes interpretations of the results of the study.

    Answer: \involvementC{} 
    
    Explanation: After executing the LLM-generated code, humans supplied the resulting data back to the LLM, which carried out the analysis, derived key findings, and produced the conclusions. Human authors’ role was limited to verifying that these AI-generated interpretations matched the actual results. 
    \item \textbf{Writing}: This includes any processes for compiling results, methods, etc. into the final paper form. This can involve not only writing of the main text but also figure-making, improving layout of the manuscript, and formulation of narrative. 

    Answer: \involvementC{} 
    
    Explanation: LLMs generated initial drafts of some sections and assisted with figure captions and language polishing. Human authors reviewed, corrected, and extended these drafts with domain-specific details and final structure to ensure accuracy and completeness. 

    \item \textbf{Observed AI Limitations}: What limitations have you found when using AI as a partner or lead author?

    Description: When asked to search for related works and generate complete \LaTeX{} entries, LLMs occasionally produced incorrect references, such as mismatched authors, inaccurate paper titles, or invalid citations. 
\end{enumerate}

\newpage

\section*{Agents4Science Paper Checklist}

\begin{enumerate}

\item {\bf Claims}
    \item[] Question: Do the main claims made in the abstract and introduction accurately reflect the paper's contributions and scope?
    \item[] Answer: \answerYes{} 
    \item[] Justification:  The abstract and introduction accurately describe the multi-agent TEAM-PHI framework, experiments with multiple LLMs, and validation via ground truth labels and human evaluation. These claims match the results in Sections 5.1–5.3 and the conclusions. 

\item {\bf Limitations}
    \item[] Question: Does the paper discuss the limitations of the work performed by the authors?
    \item[] Answer: \answerYes{} 
    \item[] Justification: Appendix C (Limitations) details dataset size constraints, potential evaluator bias, prompt sensitivity, compute costs, and fairness and regulatory adaptation issues. 
    \item[] Guidelines:
    \begin{itemize}
        \item The answer NA means that the paper has no limitation while the answer No means that the paper has limitations, but those are not discussed in the paper. 
        \item The authors are encouraged to create a separate "Limitations" section in their paper.
        \item The paper should point out any strong assumptions and how robust the results are to violations of these assumptions (e.g., independence assumptions, noiseless settings, model well-specification, asymptotic approximations only holding locally). The authors should reflect on how these assumptions might be violated in practice and what the implications would be.
        \item The authors should reflect on the scope of the claims made, e.g., if the approach was only tested on a few datasets or with a few runs. In general, empirical results often depend on implicit assumptions, which should be articulated.
        \item The authors should reflect on the factors that influence the performance of the approach. For example, a facial recognition algorithm may perform poorly when image resolution is low or images are taken in low lighting. 
        \item The authors should discuss the computational efficiency of the proposed algorithms and how they scale with dataset size.
        \item If applicable, the authors should discuss possible limitations of their approach to address problems of privacy and fairness.
        \item While the authors might fear that complete honesty about limitations might be used by reviewers as grounds for rejection, a worse outcome might be that reviewers discover limitations that aren't acknowledged in the paper. Reviewers will be specifically instructed to not penalize honesty concerning limitations.
    \end{itemize}

\item {\bf Theory assumptions and proofs}
    \item[] Question: For each theoretical result, does the paper provide the full set of assumptions and a complete (and correct) proof?
    \item[] Answer: \answerNA{} 
    \item[] Justification: The paper focuses on empirical methodology and algorithms rather than formal theorems or proofs.  
    \item[] Guidelines:
    \begin{itemize}
        \item The answer NA means that the paper does not include theoretical results. 
        \item All the theorems, formulas, and proofs in the paper should be numbered and cross-referenced.
        \item All assumptions should be clearly stated or referenced in the statement of any theorems.
        \item The proofs can either appear in the main paper or the supplemental material, but if they appear in the supplemental material, the authors are encouraged to provide a short proof sketch to provide intuition. 
    \end{itemize}

    \item {\bf Experimental result reproducibility}
    \item[] Question: Does the paper fully disclose all the information needed to reproduce the main experimental results of the paper to the extent that it affects the main claims and/or conclusions of the paper (regardless of whether the code and data are provided or not)?
    \item[] Answer: \answerYes{} 
    \item[] Justification: Sections 3–4 describe all steps of the framework, prompts, and evaluation metrics. Appendix B provides dataset description, model links, metric definitions, validation procedure, and compute resources, enabling reproducibility even with masked ground truth labels. 
    \item[] Guidelines:
    \begin{itemize}
        \item The answer NA means that the paper does not include experiments.
        \item If the paper includes experiments, a No answer to this question will not be perceived well by the reviewers: Making the paper reproducible is important.
        \item If the contribution is a dataset and/or model, the authors should describe the steps taken to make their results reproducible or verifiable. 
        \item We recognize that reproducibility may be tricky in some cases, in which case authors are welcome to describe the particular way they provide for reproducibility. In the case of closed-source models, it may be that access to the model is limited in some way (e.g., to registered users), but it should be possible for other researchers to have some path to reproducing or verifying the results.
    \end{itemize}

\item {\bf Open access to data and code}
    \item[] Question: Does the paper provide open access to the data and code, with sufficient instructions to faithfully reproduce the main experimental results, as described in supplemental material?
    \item[] Answer: \answerNo{} 
    \item[] Justification: The repository link for code is provided in the Abstract, but the real clinical note dataset cannot be shared due to PHI restrictions.   
    \item[] Guidelines:
    \begin{itemize}
        \item The answer NA means that paper does not include experiments requiring code.
        \item Please see the Agents4Science code and data submission guidelines on the conference website for more details.
        \item While we encourage the release of code and data, we understand that this might not be possible, so “No” is an acceptable answer. Papers cannot be rejected simply for not including code, unless this is central to the contribution (e.g., for a new open-source benchmark).
        \item The instructions should contain the exact command and environment needed to run to reproduce the results. 
        \item At submission time, to preserve anonymity, the authors should release anonymized versions (if applicable).
    \end{itemize}

\item {\bf Experimental setting/details}
    \item[] Question: Does the paper specify all the training and test details (e.g., data splits, hyperparameters, how they were chosen, type of optimizer, etc.) necessary to understand the results?
    \item[] Answer: \answerYes{} 
    \item[] Justification: Section 4 details dataset size, model use, and evaluation agents. Appendix B explains metrics and model links, and Section 3.4 defines aggregation and normalization procedures.  

\item {\bf Experiment statistical significance}
    \item[] Question: Does the paper report error bars suitably and correctly defined or other appropriate information about the statistical significance of the experiments?
    \item[] Answer: \answerYes{} 
    \item[] Justification: Table 8 reports mean ± standard deviation for precision, recall, and F1. Variability across evaluators and entities is discussed in Section 5. 
    \item[] Guidelines:
    \begin{itemize}
        \item The answer NA means that the paper does not include experiments.
        \item The authors should answer "Yes" if the results are accompanied by error bars, confidence intervals, or statistical significance tests, at least for the experiments that support the main claims of the paper.
        \item The factors of variability that the error bars are capturing should be clearly stated (for example, train/test split, initialization, or overall run with given experimental conditions).
    \end{itemize}

\item {\bf Experiments compute resources}
    \item[] Question: For each experiment, does the paper provide sufficient information on the computer resources (type of compute workers, memory, time of execution) needed to reproduce the experiments?
    \item[] Answer: \answerYes{} 
    \item[] Justification: Appendix B.5 lists hardware (e.g., dual NVIDIA H100 GPUs, M2 Pro 32GB), per-run times (under two hours for H100 runs), and total compute (~200 GPU-hours), and describes Azure API use for GPT models.

    \item[] Guidelines:
    \begin{itemize}
        \item The answer NA means that the paper does not include experiments.
        \item The paper should indicate the type of compute workers CPU or GPU, internal cluster, or cloud provider, including relevant memory and storage.
        \item The paper should provide the amount of compute required for each of the individual experimental runs as well as estimate the total compute. 
    \end{itemize}
    
\item {\bf Code of ethics}
    \item[] Question: Does the research conducted in the paper conform, in every respect, with the Agents4Science Code of Ethics (see conference website)?
    \item[] Answer: \answerYes{} 
    \item[] Justification: The work adheres to HIPAA privacy standards, uses de-identification methods to protect PHI, and carefully handles sensitive data as detailed in Sections 1 and 4 and Appendix C.  
    \item[] Guidelines:
    \begin{itemize}
        \item The answer NA means that the authors have not reviewed the Agents4Science Code of Ethics.
        \item If the authors answer No, they should explain the special circumstances that require a deviation from the Code of Ethics.
    \end{itemize}

\item {\bf Broader impacts}
    \item[] Question: Does the paper discuss both potential positive societal impacts and negative societal impacts of the work performed?
    \item[] Answer: \answerYes{} 
    \item[] Justification: The paper notes that the proposed framework can advance secure and large-scale reuse of clinical notes for healthcare research while protecting patient privacy, and it also warns of risks such as evaluator bias, fairness concerns, and possible misuse of de-identified data. Sections 5–6 and Appendix C.6 describe these impacts and outline mitigations like ensemble voting and human verification.  
    \item[] Guidelines:
    \begin{itemize}
        \item The answer NA means that there is no societal impact of the work performed.
        \item If the authors answer NA or No, they should explain why their work has no societal impact or why the paper does not address societal impact.
        \item Examples of negative societal impacts include potential malicious or unintended uses (e.g., disinformation, generating fake profiles, surveillance), fairness considerations, privacy considerations, and security considerations.
        \item If there are negative societal impacts, the authors could also discuss possible mitigation strategies.
    \end{itemize}

\end{enumerate}

\end{document}